\documentclass[12pt]{amsart}
\usepackage{graphicx}
\newtheorem{thm}{Theorem}

\newtheorem{lem}[thm]{Lemma}

\newcommand{\norm}[1]{\left\Vert#1\right\Vert}

\newcommand{\Real}{\mathbb R}

\newcommand{\xhat}{\hat{x}}
\renewcommand{\b}[1]{\mathbf{#1}}

\newcommand{\bU}{\b{U}}

\newcommand{\citep}[1]{(\cite{#1})}

\begin{document}

\title[Policy Iteration for Training RNNs]{Applying Policy Iteration for Training Recurrent Neural Networks \\ \vspace{3mm}Technical Report}
\author[I. Szita and A. L\H{o}rincz]{Istv\'{a}n Szita \and Andr\'{a}s L\H{o}rincz}
\address{\\ \newline Department of Information Systems  \newline E\"{o}tv\"{o}s Lor\'{a}nd University of
Sciences  \newline P\'azm\'any P\'eter s\'et\'any 1/C  \newline Budapest, Hungary H-1117\\ \newline WWW:
nipg.inf.elte.hu  \newline E-mails: \newline szityu@stella.eotvos.elte.hu \newline  andras.lorincz@elte.hu}

\maketitle

\begin{abstract} \normalsize
Recurrent neural networks are often used for learning time-series data. Based on a few assumptions we model this
learning task as a minimization problem of a nonlinear least-squares cost function. The special structure of the cost
function allows us to build a connection to reinforcement learning. We exploit this connection and derive a convergent,
policy iteration-based algorithm. Furthermore, we argue that RNN training can be fit naturally into the reinforcement
learning framework.
\end{abstract}


\section{Introduction}

\begin{figure}[h!]
\centering
\includegraphics[width=12cm]{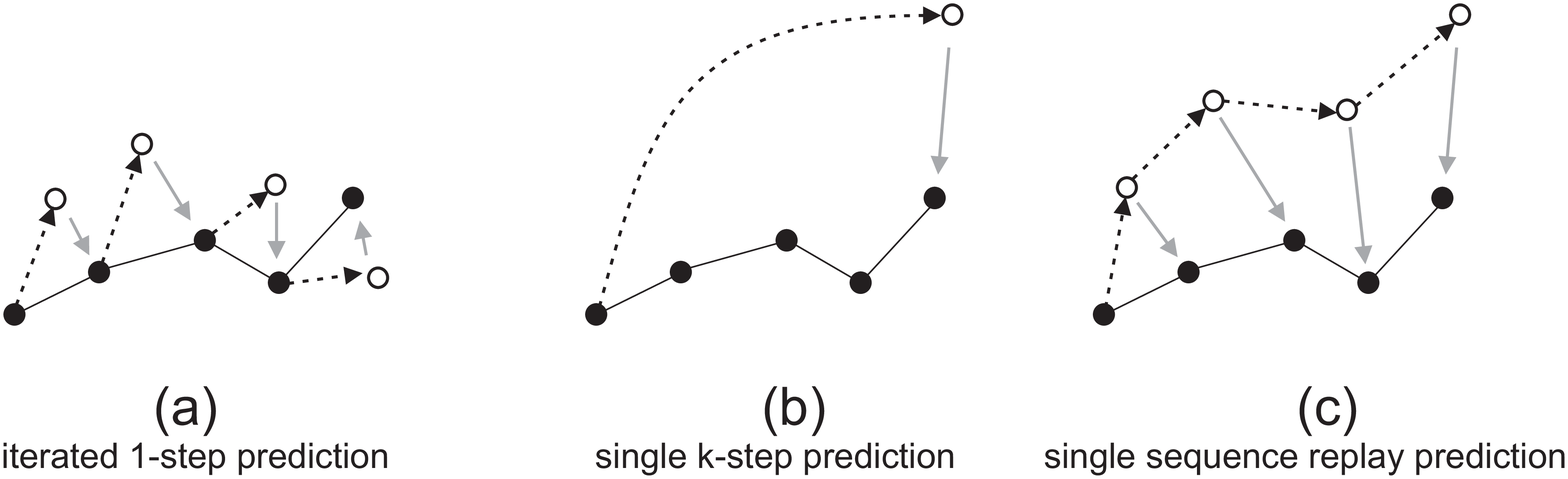}
\caption{\textbf{Comparison of time series learning methods} \newline Black dots: the original time series, dotted
arrows: predictions, gray arrows: corrections during learning.
  }\label{fig:prediction_diagram}
\end{figure}

Recurrent neural networks (RNNs) are attractive tools for the learning of time-series. However, traditional long-term
prediction methods either work as iterated 1-step methods (Fig.~\ref{fig:prediction_diagram}(a)) or by direct learning
of the $k$-step-ahead value (Fig.~\ref{fig:prediction_diagram}(b)). There is vast amount of literature on RNNs, which
is reviewed in the accompanying paper\footnote{This technical report is a supplementary material for the manuscript
`PIRANHA: Policy Iteration for Recurrent Artificial Neural Networks with Hidden Activities' written by I. Szita and A.
L{\H o}rincz. The interested reader is kindly referred to the a recent review on this topic \citep{bone02multistep} and
references therein.} In the former scheme, during the learning phase, prediction errors are computed relative to the
previous sequence values, so errors do not accumulate. However, this does not capture well the behavior in the testing
phase, when errors cannot be corrected step-by-step, so they accumulate rather quickly. The latter scheme escapes this
trap, but it needs a different estimator for each lookahead time $k$, which is far from being economic (although
implementations exist, see, e.g. \citep{duhoux01improved}). We take a novel approach: perform iterated prediction
without correction, and formulate an objective function that directly takes all the prediction errors into account (see
Fig.~\ref{fig:prediction_diagram}(c)).

The resulting objective function is a least-squares function of highly nonlinear terms, being intractable for
traditional minimization techniques. We proceed by showing that the task can be interpreted as a reinforcement learning
problem of a hypothetical agent in an abstract environment. This connection  enables the minimization of our objective
function using a version of policy iteration.

The outline of the paper is as follows. The theoretical part is made of three sections. First, we define the learning
task and derive the learning rules of our algorithm in Section~\ref{s:definitions}. In Section~\ref{s:piranha_pi} the
optimization problem is rewritten into a Policy Iteration Algorithm for Recurrent Artificial (neural) Networks with
Hidden Activities (PIRANHA).\footnote{PIRANHA can be viewed as a gradient based approach. The purpose of this technical
report is to extend the gradient based description of the accompanying paper to the framework of RL and to
\textit{provide insight} why and how PIRANHA works.} Using this novel form, we give a convergence proof of PIRANHA in
Section~\ref{s:analysis}. In Section \ref{s:discussion} we argue that considering RNN learning as a RL process is
consistent and fits well into the ``classical'' RL framework and to ongoing recent efforts in RL. The last section
summarizes our results.

\section{Definitions and Basic Concepts} \label{s:definitions}

\subsection{The Network Architecture}

We consider fully connected recurrent neural networks with $m$  input neurons and $n>m$ hidden neurons. We do not use a
separate output layer; the states of the first $m$ hidden neurons are considered as outputs. The state of each neuron
is a real number in the interval $[-1,+1]$, because neurons admit activation -- squashing -- function, which maps onto
interval $[-1,+1]$. An example is function $\sigma(z) = \tanh(z)$. The input, output and the hidden state at time $t$
are denoted by $v_t \in [-1,+1]^{m}$, $\hat{x}_{t+1} \in [-1,+1]^{m}$ and $u_t \in [-1,+1]^{n}$, respectively. No
explicit bias term is applied, instead a constant 1 might make the $(m+1)^{st}$ component of the input. Let the
recurrent, the input and the output weight matrices be denoted by $F \in \Real^{n\times n}$, $G \in \Real^{n\times
{m+1}}$, and $H \in \Real^{m\times n}$, respectively. Let $H$ be a fixed matrix projecting to the first $m$ components
of $u_t$. The dynamics of the network is as follows:
\begin{eqnarray}
  u_{t+1} &=& \sigma(Fu_{t}+Gv_{t}), \label{e:u_t+1}\\
  \hat{x}_{t+1} &=& Hu_{t+1}. \label{e:x_hat_t+1}
\end{eqnarray}
with the initial condition $u_0 = \b{0}$.

\subsection{Problem Description}

Suppose that a time series $\{x_t \in \Real^m \mid t=1,2,3,\ldots \}$ and a sequence length $T$ is given. We have to
find weight matrices $F$ and $G$ so that the network is able to replay the first $T$ elements of the sequence in
correct temporal order, i.e. for any $1<t<T$, having input $v_{t'} := x_{t'}$ for $t'\le t$ and generating $v_{t'} :=
\hat{x}_{t'}$ for $t < t' \le T$, the total squared error $\sum_{t'=t}^{T} \norm{\hat{x}_{t'}-x_{t'}}^2$ is small.

The sum of the least-square errors for all  possible $t'$ values characterizes the replay capability of the network
with the network weights given. However, this cost function may have pitfalls for ordinary gradient-based methods,
because little changes in weights may considerably influence the output many steps ahead. Sensitivity may become
crucial under this condition.

Our key observation to the solution is that the same problem emerges in reinforcement learning (RL). We shall
reformulate the learning task to highlight this connection and we shall apply RL algorithms to optimize our problem.

\subsection{Constructing an Appropriate Cost Function}

The traditional approach would be to search for weight matrices $F$ and $G$ that minimize the reconstruction error
\begin{equation} \label{e:J_naive}
  \sum_{t=0}^T \norm{\xhat_t - x_t}^2 = \sum_{t=0}^T \norm{Hu_t - x_t}^2
\end{equation}
subject to the constraints (\ref{e:u_t+1})--(\ref{e:x_hat_t+1}). It is well known that such recurrent network
optimization tasks are hard, because the objective function has many local minima and is often ill-conditioned. Let us
investigate some possible reasons: consider the case when all the predicted output values $\xhat_t$ are correct, except
for a single time step $t_0$. The hidden state $u_{t_0}$ for this time step is also bad, and we can modify it only by
modifying the weights $F$ and $G$. However, this modification is likely to compromise all the other errors, resulting
in a total increase in the objective value. The reason for this phenomenon is that the hidden states for different time
steps are very strongly coupled by Eq.~(\ref{e:u_t+1}).

The underlying idea of our approach is that if this coupling is relaxed, the resulting error surface may become
smoother. Naturally, in the end we have to enforce Eq,~(\ref{e:u_t+1}), but first let us consider a different set of
constraints. Suppose that the sequence $\bU = (u_0, u_1, u_2, \ldots)$ is an arbitrary state sequence, not necessarily
belonging to an RNN. When can we say that matrices $F$, $G$ and a state $u_t$ in the sequence is a good predictor? The
one-step prediction from $u_t$ is
\begin{equation}
  \xhat_{t\to(t+1)} = H\sigma(Fu_t+Gx_t),
\end{equation}
for two steps it is
\begin{eqnarray}
  \xhat_{t\to(t+2)} &=& H\sigma(F\sigma(Fu_t+Gx_t)+G\xhat_{t\to(t+1)}) \\
                &=& H\sigma\left( (F+GH)\sigma(Fu_t+Gx_t)\right), \nonumber
\end{eqnarray}
and so on. Let us introduce the notation
\begin{equation} \label{e:s^1}
   s^{(1)}_{t,(F,G)} (u) := \sigma(F u + G x_{t})
\end{equation}
for the effect of $(F,G)$ on a single RNN state $u\in\Real^n$ and
\begin{equation} \label{e:s^k}
   s^{(k)}_{t,(F,G)} (u) := \sigma\left((F +GH)s^{(k-1)}_{t,(F,G)} (u) \right)
\end{equation}
to describe the effect of applying $(F,G)$ $k \,\, (>1)$ times on state $u$. For later use we also define $s^{(0)}$ as
the identity function.

Using these notations, the $k$-step prediction error of state $u_t$ is
\begin{equation}
 \left(s^{(k)}_{t,(F,G)} (u_t) - x_{t+k}\right).
\end{equation}
(For the sake of notational simplicity, we assume that the input sequence $x_t$ is defined for $t>T$ as well.) The
overall cost of state $u_t$ should be the the sum of the $k$-step prediction error norms, for all time instants and for
all $k$s. To ensure convergence, we use a geometrically decaying weight sequence with decay rate $\gamma$. So the total
cost of the state sequence $\bU$ is
\begin{equation} \label{e:J}
   J(\bU, F,G) = \sum_{k=0}^{\infty} \sum_{t=1}^{T} \gamma^k \norm{H \bigl( s_{t,(F,G)}^{(k)} (u_t)\bigr) - x_{t+k}}^2
\end{equation}

Note that if a set of weights $(F,G)$ minimizes Eq.~(\ref{e:J_naive}) subject to Eq.~(\ref{e:u_t+1}), it also minimizes
Eq.~(\ref{e:J}), and technical assumptions can ensure that the converse also holds. Note also that cost function in
Eq.~(\ref{e:J}) has a large number of partial sequences and so, it has an enormous number of adjustable quantities.
This gives one great freedom in accomplishing the minimization. However, there is a price to pay: minimization is
ill-posed, because the number of parameters is much larger than the number of data points. We restrict the choices to
the parameters of the original problem, but in a way which is different from Eq.~(\ref{e:J_naive}), and which reflects
the structure of the sequence replay problem better.

To this end, note that if a particular state $u_t$ is a good predictor at time step $t$ using weights $(F,G)$, then
$\sigma(Fu_t + Gx_t)$ is a good predictor at time step $t+1$, following from the structure of the cost function. Let
\begin{eqnarray}
   S_{(F,G)} \bU &:=& (u'_0, u'_1, \ldots, u'_T), \quad\textrm{where} \\
    && u'_0 = \b{0}, \nonumber \\
    && u'_{t+1} = \sigma(Fu_t + Gx_t) \quad\textrm{for $0\le t < T$}. \nonumber
\end{eqnarray}
Using this notation the above statement says that if $\bU$ is a good predictor with $(F,G)$, $S_{(F,G)} \bU$ is also a
good predictor. This argument can be applied iteratively, yielding that $S_{(F,G)}^k \bU$ is a good predictor for
$k>1$. However, it is easy to see that for $k>T$ $S_{(F,G)}^k \bU$ is equal to the state sequence generated by
(\ref{e:u_t+1}), and will be denoted by $\bU_{(F,G)}$:
\begin{eqnarray} \label{e:U_FG}
   \bU_{(F,G)} &:=& (u_0, u_1, \ldots, u_T), \quad\textrm{where} \\
    && u_0 = \b{0}, \nonumber \\
    && u_{t+1} = \sigma(Fu_t + Gx_t) \quad\textrm{for $0\le t < T$}. \nonumber
\end{eqnarray}

This justifies the following improvement scheme:
\begin{itemize}
  \item fix $\bU$, $F_0$, $G_0$
  \item find $(F,G)$ for which $J(S_{(F,G)}\bU, F_0, G_0)$ is small
  \item continue with $\bU :=\bU_{(F,G)}$, $F_0:=F$ and $G_0 = G$.
\end{itemize}

In Section~\ref{s:analysis} we will formally prove the correctness of the method, but before doing so, we elaborate on
some of the details.

First of all, note that the number of adjustable parameters is the same as in the original method, but the parameters
$F$ and $G$ play a different role: they are chosen so that they minimize the multi-step prediction errors, and relation
(\ref{e:u_t+1}) is part of the full optimization problem.\footnote{Several numerical simulations showed the step $\bU
:=\bU_{(F,G)}$ could be substituted by $\bU :=S_{(F,G)} \bU$ or $\bU :=S^k_{(F,G)} \bU$ for any $k$ and under such
conditions, constraint (\ref{e:u_t+1}) does not appear at all. However, theoretical analysis is easier for the
definition, which includes Eq.~(\ref{e:u_t+1}) and this definition will be used here.}

\subsection{Computing the Gradient}

In this subsection we work out the details of the above scheme.

Let us denote the state sequence at iteration $i$ by $\bU_i$, and the weight matrices by $F_i$ and $G_i$, respectively.
We would like to find $F$ and $G$ so that
\begin{eqnarray} \label{e:J'}
  J'(F,G) &:=& J(S_{(F,G)}\bU_i, F_i, G_i) = \\
    &=& \sum_{t=1}^T \sum_{k=1}^\infty \gamma^k  \norm{H \bigl( s_{{t+1},(F_i,G_i)}^{(k-1)} s_{t,(F,G)}^{(1)} (u_t)\bigr) -
    x_{t+k}}^2 \nonumber\\
    &:=& \sum_{t=1}^T \sum_{k=1}^\infty \gamma^k  \norm{e_{t,k}}^2 \nonumber
\end{eqnarray}
is minimized, that is, we are minimizing the cost by propagating all states $u_t$ by $(F,G)$ once, and then by
propagating with $(F_i, G_i)$ further.

In order to do this, we have to compute the gradient of the cost function with respect to the weights. For any $1\le
a,b,c \le n$ and $1\le d \le m+1$, the gradients are given by
\begin{eqnarray}  \label{e:dJ'/dF}
  \Bigl( \frac{\partial {J}'\ }{\partial F_{ab}} \Bigr)(F_i,G_i) = \sum_{t=1}^T \sum_{k=1}^\infty \gamma^k \left( e_{t,k}  \right)^\top
     H [m(k)]_a [u_{t}]_{b}
\end{eqnarray}
and
\begin{eqnarray}
  \Bigl( \frac{\partial {J}'\ }{\partial G_{cd}} \Bigr)(F_i,G_i) = \sum_{t=1}^T \sum_{k=1}^\infty \gamma^k \left( e_{t,k}
\right)^\top
     H [m(k)]_c \left[\begin{array}{c}  x_{t+k} \\  1 \end{array} \right]_{d}, \label{e:dJ'/dG}
\end{eqnarray}
where
\begin{eqnarray}
 m(k) := \prod_{j=1}^k \sigma' \left( [s_{t,(F_i,G_i)}^{(j-1)} u_t] \right) F_i, \label{e:m_k}
\end{eqnarray}
$(.)^\top$ denotes transposition and $[v]_b$ denotes the $b^{th}$ component of vector $v$.

If we can ensure that the terms $m(k)$ remain bounded, then the above sums are convergent, because all terms can be
bounded by $C\cdot \gamma^k$. This means that if $K$ is a sufficiently large positive number, then the gradients of
\begin{equation} \label{e:J'_approx}
  \hat{J}'(F,G) := \sum_{t=1}^T \sum_{k=1}^K \gamma^k  \norm{e_{t,k}}^2
\end{equation}
will be arbitrarily close to (\ref{e:dJ'/dF}) and (\ref{e:dJ'/dG}), so we can use it as an approximation.

Using the above formulae, we can obtain weight matrices better than $(F_i,G_i)$ by taking a gradient descent step. The
description of the algorithm is therefore complete. We have summarized it in Fig.~\ref{alg1}. In the next section we
will show that the algorithm is can be seen as policy iteration of an RL problem, which justifies its name,
\emph{Policy Iteration for Recurrent Artificial (neural) Networks with Hidden Activities}, or \emph{PIRANHA} for short.

\begin{figure}[h]
 \hrule \vskip1pt \hrule \vskip1mm
\begin{tabbing}
xxxxx \= xxx \= xxx \= xxx \= xxxxxxxxxxx \= \kill
\> input: $\alpha$, $K$, $T$, $(x_1, \ldots, x_T)$; \\
\> initialize $F_0, G_0$; $i:=0$; \\
\> for $i:=1$ to $max\_iter$ \\
\> \> $\bU_i := \bU_{(F_i,G_i)}$;\\
\> \> for $t:=0$ to $T$ \\
\> \> \> $m_t(1) := F_i$; \\
\> \> \> for $k=1...(K-1)$, \\
\> \> \> \> $m_t(k+1):= m_t(k) \cdot \sigma'\left( [s_{t,(F_i,G_i)}^{(k)} u_t] \right) F_i$; \\
\> \> \> \> $e_{t,k}:=H \bigl( s_{{t+1},(F_i,G_i)}^{(k-1)} s_{t,(F,G)}^{(1)} (u_t)\bigr) - x_{t+k}$; \\
\> \> $[\Delta F]_{ab} := \sum_{t=1}^T \sum_{k=1}^K \gamma^k \left(  e_{t,k}  \right)^\top
     H [m_t(k)]_a [u_{t}]_{b}$; \\
\> \> $[\Delta G]_{cd} := \sum_{t=1}^T \sum_{k=1}^K \gamma^k \left( e_{t,k} \right)^\top H
      [m_t(k)]_c [x_{t+k};1]_d$; \\
\> \> $F_{i+1} = F_i - \alpha_i \cdot \Delta F$; \\
\> \> $G_{i+1} = G_i - \alpha_i \cdot \Delta G$; \\
\> \> $i:=i+1$; \\
\> end
\end{tabbing}
 \hrule \vskip1pt \hrule
 \caption{Pseudo-code of the PIRANHA algorithm} \label{alg1}
\end{figure}

\section{Interpreting PIRANHA as Policy Iteration} \label{s:piranha_pi}

Notice that the cost function (\ref{e:J}) is formally very similar to the cost function of a reinforcement learning
problem. Furthermore, the proposed algorithmic solution is very much like policy iteration, a widely used algorithm for
solving RL problems. Although this similarity is only formal, as in our case uncertain transitions are not considered,
the policy iteration formalism can be matched perfectly, and this gives us valuable insight how PIRANHA works.
Furthermore, the similarity enables us to prove that under appropriate conditions the algorithm is convergent.

Firstly, we give a brief overview of the reinforcement learning framework and policy iteration. Next we reformulate the
sequence learning problem as a special case of RL, and point out some important differences concerning traditional RL
problems. Finally, using the policy iteration formalism, we present our main result about the convergence of PIRANHA.

\subsection{The Reinforcement Learning Framework and Policy Iteration} \label{s:RL}

RL deals with agents that make decisions, i.e., selects actions in a stochastic environment. The state of the
environment is influenced by the decisions of the agent. From the point of view of the agent, the actual state is
(fully or partially) observed, actions are selected and state-dependent immediate costs are received. RL aims to
minimize the total cumulated cost by finding the optimal decisions (\emph{optimal policy}) for the agent.

In most cases, RL problems are treated as Markov decision processes, i.e., states are fully observable and rewards and
successor states depend only on the current state and action but not on the history. For the sake of simplicity, we
also assume that costs are bounded and deterministic, and depend only on the current state. Let us denote the state
space and the actions of the agent by $\mathcal{S}$ and $\mathcal{A}$, respectively. The dynamics of the environment is
characterized by the transition probability function $P: \mathcal{S}\times\mathcal{A}\times\mathcal{S} \to [0,1]$ and
the immediate cost function $c: \mathcal{S}\to \Real$. A deterministic policy $\pi:\mathcal{S} \to \mathcal{A}$ of the
agent is a mapping from states to actions. Future costs to be cumulated may be discounted by some factor
$0\le\gamma\le1$, so the expected cost of a state $s_0$ is
\begin{equation} \label{e:J_def}
  J^\pi(s_0) = E(c(s_0) + \gamma c(s_1) + \gamma^2 c(s_2) + \ldots ),
\end{equation}
where $E(.)$ denotes expectation, and for each $t\ge 0$, $a_t = \pi(s_t)$ and the system arrives at $s_{t+1}$ with
probability $P(s_t,a_t,s_{t+1})$. The task is to find an optimal policy $\pi^*$, for which $J^{\pi^*}(s) \le
J^{\pi}(s)$ for any state $s$ and any policy $\pi$.

\emph{Policy iteration} approaches the optimal policy by a two-phase iteration procedure: at each iteration $i$, the
current policy $\pi_i$ is first \emph{evaluated}, i.e. $J^{\pi_i}(s)$ is computed (or approximated), usually by letting
the agent to take many steps using $\pi_i$ and processing the experienced costs. The other phase is called \emph{policy
improvement}. In this phase the policy is modified by using the inequality
\begin{eqnarray}
  J^{\pi_i}(s) &=& c(s) + \gamma \sum_{s'} P(s,\pi(s),s') J^{\pi_i}(s')\\
   &\ge& \min_a \Bigl(  c(s) + \gamma \sum_{s'} P(s,a,s') J^{\pi_i}(s') \Bigr),
\end{eqnarray}
which implies that
\begin{equation} \label{e:policy_improvement}
 \pi_{i+1}(s) := \arg\min_a \Bigl(  c(s) + \gamma \sum_{s'} P(s,a,s') J^{\pi_i}(s') \Bigr)
\end{equation}
is an improvement over $\pi_i$.\footnote{Note that for policy improvement, it is not necessary to choose the minimum in
the equation.} The improvement can be gradual: there is no need to find the minimum, but it is sufficient if a partial
and/or approximate step is made in that direction.

If transition probabilities, policies and cost function values are stored in a lookup-table, with separate entries for
each different argument, policy iteration converges to an optimal policy under appropriate conditions. This is true
even if approximations are used in either step or $P$ and $c$ is not known in advance but have to be estimated from the
agent-environment interaction. For details about the conditions of the various convergence results, see
\citep{bertsekas96neuro-dynamic,sutton98reinforcement}.

In some cases, the construction of the lookup-table is not feasible, e.g., when the state and/or the action spaces are
continuous. Then one needs to revert to function approximation methods. It has been shown, however, that policy
iteration for function approximators can be divergent even for the simplest cases \citep{bertsekas96neuro-dynamic} and,
in turn, the proof of convergence becomes a central issue.

\subsection{Fitting PIRANHA into the RL Framework}

\subsubsection{The Agent and its Environment}
Let us define a hypothetical agent that observes the state sequence $\bU$, and based on that, chooses a set of weights
$(F,G)$, which is used for prediction. Formally this corresponds to $\mathcal{S} = \Real^{n\times T}$, so that a state
$\bU\in\mathcal{S}$ of the agent is the whole sequence of hidden states of the RNN: $\bU=(u_1, u_2, \ldots, u_T)$,
while the action space will be $\mathcal{A} := \Real^{n\cdot n + n\cdot(m+1)}$. The environment of this hypothetical
agent is quite simple: if the agent makes an action $(F,G)$ in state $\bU$, then the environment transfers it
deterministically to the successor state $S_{(F,G)}\bU$.

In a general-case RL problem, a policy of the agent is a mapping $\pi: \mathcal{S} \to \mathcal{A}$. However, in our
problem, time independent weight matrices ($F$ and $G$) are searched for, therefore we restrict the mapping to constant
policies. Such policies $\pi_{(F,G)}$ execute the fixed action $(F,G)$ regardless of the current state. For a policy
$\pi \equiv (F,G)$ we will also use the notation $S_\pi(.)$ instead of $S_{(F,G)}(.)$.

\subsubsection{Costs}
Let $\bU = (u_1, u_2, ..., u_T)$ be an arbitrary state. Its immediate cost $c(\bU)$ is defined as the error of
reconstructing the input series $x_t$:
\begin{equation}\label{e:J_naive_2}
c(\bU) := \sum_{t=1}^{T} \norm{Hu_t - x_t}^2.
\end{equation}
Thus, by (\ref{e:J_def}), the total discounted cost of
$\bU$ using policy $\pi\equiv(F,G)$ and discount factor $0\le\gamma < 1$ is
\begin{eqnarray} \label{e:J_RL}
  J^\pi(\bU) &:=& c(\bU) + \gamma c(S_\pi \bU) + \gamma^2 c(S^2_\pi \bU) + ... \\
        &=& \sum_{k=0}^\infty \sum_{t=1}^{T} \gamma^k \norm{H \bigl( s_{t,(F,G)}^{(k)} (u_t)\bigr) - x_{t+k}}^2,
\end{eqnarray}
which is the same cost function as (\ref{e:J}). This cost function reverts to the cost function of Eq.~\ref{e:J_naive}
(=Eq.~\ref{e:J_naive_2}) for $\gamma=0$, but then policy iteration reduces to the iterative 1-step method.

\subsubsection{Policy Evaluation}
At iteration $i>0$, we have a policy $\pi_i = \pi_{(F_i,G_i)}$ to be evaluated. The evaluation starts by taking many
steps using $\pi_i$, which (by (\ref{e:U_FG})) guarantees that the agent is in state $\bU_{(F_i,G_i)}$. Therefore we
have to evaluate the cost function in a single state $\bU_{(F_i,G_i)}$ only, which -- in contrast to traditional policy
iteration -- can be done directly.

\subsubsection{Policy Improvement}
Policy improvement will be accomplished only for the single state $\bU_{(F_i,G_i)}$, because this is the only state
where the full cost is known. As a further deviation from the basic policy improvement strategy
(\ref{e:policy_improvement}), we take only a small step towards
\begin{eqnarray}
  &&\arg\min_{(F,G)} \left( c(\bU_{(F_i,G_i)}) + \gamma \hat{J}^{\pi_i}(S_{(F,G)} \bU_{(F_i,G_i)}) \right) := \\
  &&\arg\min_{(F,G)} \hat{J}'(F,G),
\end{eqnarray}
and improve the policy gradually:
\begin{equation} \label{e:pi_improvement}
  \pi_{i+1}(\bU_{(F_i,G_i)}) = (F_{i+1},G_{i+1}):= (F_{i},G_{i}) - \alpha_i \nabla \hat{J}'(F_i,G_i),
\end{equation}
where $\alpha_i$ is the learning rate for iteration $i$, and $\nabla$ denotes the gradient with respect to components
of $(F,G)$.

Iterating these two steps completes policy iteration, and as it can be easily seen, it is identical to the PIRANHA
algorithm described in Fig.~\ref{alg1}.

\subsubsection{Differences from Standard Policy Iteration}

There are several important differences between standard policy iteration algorithm and PIRANHA. Firstly, in the policy
improvement step we modified the policy so that only the cost of a \emph{single state}, $\bU_i$ is considered. However,
changing the policy in $\bU_i$ changes the costs of every other state as well, and we have no guarantee that the change
is really an improvement for all states (in fact, it can be shown that there will be states for which the cost
increases). Thus, we cannot apply any of the convergence results of (approximate) policy iteration algorithms, because
they all require improvement over the whole state space. Fortunately, $J^{\pi_i}(\bU_i) \ge J^{\pi_{i+1}}(\bU_{i+1})$
still holds, but we need to prove this by taking into account the specific structure of the cost function.

\section{Analysis} \label{s:analysis}

\subsection{The Finite-$K$ Approximation}

The gradients of (\ref{e:J'}) are infinite sums, which are approximated by finite sums up to the $K^{th}$ term. The
first question we have to answer is whether this approximation is feasible. The following lemma shows that for
sufficiently small discount factor $\gamma$, the gradients (\ref{e:dJ'/dF}) and (\ref{e:dJ'/dG}) are convergent,
therefore for sufficiently large $K$, the gradients of (\ref{e:J'}) will be approximated well by (\ref{e:J'_approx}).

\begin{lem} \label{thm:conv_approx}
Suppose that (a) For all $i$, $\norm{F_i}_\infty \le 1/\sqrt\gamma$, (b) The slope of the sigmoid function is at most
1, i.e. $\sup_z \sigma'(z) \le 1$. Then for any $\epsilon_0 > 0$ there exists a $K$ so that if (\ref{e:J'_approx}) is
used for calculating the gradient, the difference is less than $\epsilon_0$.
\end{lem}
\begin{proof}

The difference of derivatives with respect to some component $F_{ab}$ is
\begin{eqnarray}
 && \left| \left( \frac{\partial J'\ }{\partial F_{ab}} \right)(F_i,G_i) - \left( \frac{\partial \hat J'\ }{\partial F_{ab}}
 \right)(F_i,G_i)\right| = \\
  && = \left| \sum_{t=1}^T \sum_{k=K+1}^\infty \gamma^k \left(  e_{t,k}  \right)^\top H [m(k)]_a [u_{t}]_{b}  \right| \le\\
  &&\le \sum_{k=K+1}^\infty \sum_{t=1}^T \gamma^k \norm{ e_{t,k}}_\infty \norm{H}_\infty \norm{ m(k)}_\infty \norm{
  u_{t}}_\infty
\end{eqnarray}
But $u_t$ is an inner state of a neural network, so $\norm{u_{t}}_\infty \le 1$, and similarly, $\norm{ e_{t,k}}_\infty
\le 2$, and $\norm{H}_\infty =1$ because it is a projection matrix. Using (\ref{e:m_k}) and Assumption (b), the  norm
of $m(k)$ can be bounded from above by
\begin{equation}
  \norm{m(k)}_\infty \le \norm{F_i}_\infty^k \le \gamma^{-k/2},
\end{equation}
so the difference of the exact and approximate derivatives can be bounded as
\begin{eqnarray}
  &&\sum_{k=K+1}^\infty \sum_{t=1}^T \gamma^k \cdot 2 \cdot 1 \cdot \gamma^{-k/2} \cdot 1 = \sum_{k=K+1}^\infty 2T
  \gamma^{k/2} =\\
  && = \frac{2T \gamma^{(K+1)/2}}{1-\gamma^{1/2}} := C_0 \gamma^{K/2},
\end{eqnarray}
which is can be made less than $\epsilon_0$ for sufficiently large $K$. One may use (\ref{e:dJ'/dG}) to show that the
same bound applies for the differences of derivatives with respect to elements of $G$.
\end{proof}

\subsection{Convergence proof}

We should not that proving the convergence of the algorithm of Table~\ref{alg1} is easy, because it is a gradient
descent method, therefore, as it is well known, it converges to a (local) minimum if the step sizes $\alpha_i$ are
sufficiently small.

However, the algorithm offers new possibilities within the framework of reinforcement learning as we shall discuss it
later. We give an alternative proof that is somewhat more complicated, but exploits the policy iteration reformulation.
This derivation reflects the mechanism and the potentials of the algorithm better.

\section{The proof of the Convergence Theorem}

First, suppose that $J$ and $J'$ can be computed exactly. We proceed by a series of lemmas:

Let $\pi_i = (F_i, G_i)$ be the policy at iteration $i$, $\bU_{\pi_i} = (u_0, u_1, \ldots, u_T)$, furthermore, let the
gradient of $J'$ be denoted by $\Delta \pi = (\Delta F, \Delta G)$ as above.

\begin{lem}\label{lem:conv1}
For $\Delta \pi \not\equiv 0$ there exists a number $\alpha_0 > 0$ such that for all $0< \alpha \le \alpha_0$,
\begin{equation}
  J^{\pi_i + \alpha \Delta \pi}(\bU_{\pi_i}) < J^{\pi_i}(\bU_{\pi_i}).
\end{equation}
\end{lem}
\begin{proof}
By definition, $\Delta\pi$ is the steepest descent direction of $c(\bU_{\pi_i}) + J^{\pi_i}(S_{\pi}\bU_{\pi_i})$, and
it is nonzero, therefore for sufficiently small $\beta_0$, for all $0 < \beta \le \beta_0$,
\begin{eqnarray}
  &&c(\bU_{\pi_i}) + \gamma J^{\pi_i}(S_{\pi_i+\beta \Delta\pi}\bU_{\pi_i}) < \\
  && < c(\bU_{\pi_i}) + \gamma J^{\pi_i}(S_{\pi_i}\bU_{\pi_i}) =
  J^{\pi_i}(\bU_{\pi_i}).
\end{eqnarray}

Let
\begin{eqnarray*}
 \mathcal{U} &:=& \{ \bU \in \Real^{n\times(T+1)} \mid  c(\bU) + \gamma J^{\pi_i}(S_{\pi_i+\beta \Delta\pi}\bU) < J^{\pi_i}(\bU) \textrm{ for all } \beta\le \beta_0\}
\end{eqnarray*}
be the set of states for which $\pi_i+\beta_0 \Delta\pi$ is an improvement over $\pi_i$. This is an open set, because
$c(\bU) + \gamma J^{\pi_i}(S_{\pi_i+\beta_0 \Delta\pi}\bU) - J^{\pi_i}(\bU)$ is a continuous function of $\bU$.
Furthermore, $\bU_{\pi_i} \in \mathcal{U}$, so it also contains a ball with some positive radius $r$ centered on
$\bU_{\pi_i}$.

Note that $\bU_{\pi_i} = S_{\pi_i} \bU_{\pi_i}$ and the operator $S_{\pi_i + \beta \Delta\pi}$ is continuous in
$\beta$, so there exists a $\beta_1$ so that $\norm{\bU_{\pi_i} - S_{\pi_i + \beta_1 \Delta\pi} \bU_{\pi_i}} \le r$,
which means that $S_{\pi_i + \beta_1 \Delta\pi} \bU_{\pi_i} \in \mathcal{U}$, and in general, for all $k$ there exists
a $\beta_{k}$ so that $S_{\pi_i + \beta_{k} \Delta\pi}^k \bU_{\pi_i} \in \mathcal{U}$. Let
\begin{equation}
  \alpha_0 := \min(\beta_0, \beta_1, \ldots, \beta_{T-1}).
\end{equation}
Then for all $0\le k < T$ and $0 < \alpha <\alpha_0$
\begin{equation}
 c(S_{\pi_i+\alpha \Delta\pi}^{k} \bU_{\pi_i}) + \gamma J^{\pi_i}(S_{\pi_i+\alpha \Delta\pi}^{k+1} \bU_{\pi_i})
   < J^{\pi_i}(S_{\pi_i+\alpha \Delta\pi}^{k} \bU_{\pi_i})
\end{equation}
by the definition of $\mathcal{U}$. Adding these inequalities with weights $\gamma^k$ gives
\begin{eqnarray}
  &&c(\bU_{\pi_i}) + \gamma c(S_{\pi_i+\alpha \Delta\pi}\bU_{\pi_i}) + \ldots + \gamma^{T-1} c(S_{\pi_i+\alpha \Delta\pi}^{T-1}\bU_{\pi_i})
  +\\
  && + \gamma^T J^{\pi_i}(S_{\pi_i+\alpha \Delta\pi}^{T} \bU_{\pi_i}) < J^{\pi_i}(\bU_{\pi_i}).
\end{eqnarray}
The LHS is exactly the cost of taking $T$ steps using $\pi_i+\alpha \Delta\pi$, and then using $\pi_i$. Noting that
this is the same as following $\pi_i+\alpha \Delta\pi$ all the way, so we get
\begin{eqnarray}
  J^{\pi_i+\alpha\Delta\pi}(\bU_{\pi_i}) < J^{\pi_i}(\bU_{\pi_i}),
\end{eqnarray}
which was to be shown.
\end{proof}

\begin{lem}\label{lem:conv2}
For $\Delta \pi \not\equiv 0$ there exists a number $\alpha_1 > 0$ such that for all $0< \alpha \le \alpha_1$,
\begin{equation}
  J^{\pi_i + \alpha \Delta \pi}(\bU_{\pi_i + \alpha \Delta \pi}) < J^{\pi_i}(\bU_{\pi_i}).
\end{equation}
\end{lem}
\begin{proof}
The proof proceeds similarly as the proof of Lemma~\ref{lem:conv1}: Let
\begin{eqnarray}
 \mathcal{U}' : = \{ \bU \in \Real^{n\times(T+1)} \mid J^{\pi_i + \alpha \Delta \pi}(\bU) < J^{\pi_i}(\bU_{\pi_i}) \\ \textrm{ for all } \alpha\le
 \alpha_0\}\nonumber
\end{eqnarray}
This set is open as well because of the continuity of $J$, and by Lemma~\ref{lem:conv1}, it contains $\bU_{\pi_i}$, and
thus contains also a ball centered around $\bU_{\pi_i}$. This means that for sufficiently small $\beta_0$, $S_{\pi_i +
\beta \Delta \pi} \bU_{\pi_i} \in \mathcal{U}'$ for all $0< \beta < \beta_0$.

Setting $\alpha_1 := \min(\alpha_0, \beta_0)$ proves the statement of the lemma.
\end{proof}

These lemmas enable us to prove convergence for the exact cost function case:

\begin{thm}\label{thm:conv}
If the learning rates $\alpha_i$ are determined by Lemma~\ref{lem:conv2}, PIRANHA converges to a policy $\bar\pi$ that
is a (local) minimum of $J^\pi(\bU_\pi)$.
\end{thm}
\begin{proof}
As Lemma~\ref{lem:conv2} states, $J^\pi(\bU_\pi)$ is monotonously decreasing, thus it is necessarily convergent, which
also means that the policy series $\pi_i$ converges to some $\bar{\pi} = \lim_{i\to\infty} \pi_i$. By taking the limit
of $\pi_{i+1} = \pi_i + \alpha_i \Delta \pi$ we get by applying lemma~\ref{lem:conv2} to $\bar\pi$ that $\bar\alpha
\Delta\bar\pi = 0$. This means that either $\Delta\bar\pi = 0$ or $\bar\alpha=0$. However, if $\Delta\bar\pi \neq 0$,
Lemmas~\ref{lem:conv1} and \ref{lem:conv2} guarantee the existence of a positive learning rate $\alpha$, so the
gradient necessarily vanishes.
\end{proof}

Now suppose that we do not compute the exact $J'$ for computing the gradient, but use the approximation
\begin{eqnarray} \label{e:J'_hat}
  \bar J'(F,G) &=& c(\bU_i) + \sum_{t=1}^T \sum_{k=1}^K \gamma^k  \norm{e_{t,k}}^2
\end{eqnarray}
instead with some fixed lookahead parameter $K$. The following lemma is a rephrasing of lemma~\ref{thm:conv_approx},
and shows that we can get arbitrarily close to the minimum if $K$ is sufficiently large and the norm of recurrent
weights does not grow much beyond 1.

To this end, let us first define the norm of a policy $\pi$: recall that $\pi$ is the concatenation of a $n \times n$
and an $n \times (m+1)$ real-valued matrix. If we regard policies as $n(n+m+1)$ dimensional vectors, then we can define
the scalar product $\langle\pi_1,\pi_2\rangle$ of two policies as normal scalar products in Euclidean space, and the
norm of policy $\pi$ as $\norm{\pi} = \langle\pi,\pi\rangle^{1/2}$. The norm of matrices and vectors is defined as
usual; $\norm{.}_\infty$ denotes the maximum-norm.

\begin{thm}
Suppose that (a) For all $i$, $\norm{F_i}_\infty \le 1/\sqrt\gamma$, (b) The slope of the sigmoid function is at most
1, i.e. $\sup_z \sigma'(z) \le 1$. Then for any $\epsilon_0 > 0$ there exists a $K$ so that if (\ref{e:J'_hat}) is used
for calculating the gradient $\hat\pi_i$, then $\lim\sup_{i\to\infty} \norm{\Delta\hat\pi_i} \le \epsilon_0$.
\end{thm}
\begin{proof}
For the proof we will exploit the fact that gradient descent is convergent with approximating the direction of the
steepest descent as long as actual direction of descent and the direction of steepest descent enclose an acute angle,
i.e. as long as the scalar product is positive. In our case this means that we have to prove that at each iteration,
$\langle \Delta\pi, \Delta\hat\pi \rangle > 0$. We begin by bounding $\norm{\Delta\pi - \Delta\hat\pi}$.

Lemma~\ref{thm:conv_approx} states that he difference of the exact and approximate derivatives can be bounded by
\begin{eqnarray}
 && \left| \left( \frac{\partial J'\ }{\partial F_{ab}} \right)(F_i,G_i) - \left( \frac{\partial \hat J'\ }{\partial F_{ab}}
 \right)(F_i,G_i)\right| \le \\
  && \le \frac{2T \gamma^{(K+1)/2}}{1-\gamma^{1/2}} := C_0 \gamma^{K/2},
\end{eqnarray}
where $C_0$ is a constant independent of $K$, and the same holds for the differences of derivatives with respect to
elements of $G$.

Therefore the norm of the gradient error $\Delta\pi - \Delta\hat\pi$ can be bounded from above as
\begin{equation}
  \norm{\Delta\pi - \Delta\hat\pi} \le \sqrt{n(n+m+1)} C_0 \gamma^{K/2} := C_1 \gamma^{K/2}.
\end{equation}
This can be made smaller than any fixed $\epsilon_0> 0$ if $K > \frac{2\log(\epsilon_0/C_1)}{\log \gamma}$.

Applying this result to the scalar product yields
\begin{eqnarray}
  \langle \Delta\pi, \Delta\hat\pi \rangle &=& \langle
    \Delta\pi,\Delta\pi \rangle + \langle \Delta\pi, \Delta\hat\pi - \Delta\pi\rangle \\
  &\ge& \norm{\Delta\pi}^2 - \norm{\Delta\pi} \norm{\Delta\pi - \Delta\hat\pi} \\
  &\ge& \norm{\Delta\pi} ( \norm{\Delta\pi} - \epsilon_0),
\end{eqnarray}
which is greater then zero if $\norm{\Delta\pi} > \epsilon_0$, but this holds by the assumption of the theorem.
\end{proof}

\noindent\emph{Remark.} Assumption (a) requires that the recurrent weights remain relatively  small (the norm of $F$
does not grow much beyond 1), thus avoiding chaotic behavior. Although this constraint cannot be verified \emph{a
priori}, it can be enforced by either choosing small $\gamma$ values or by applying weight regularization technique.

\section{Discussion} \label{s:discussion}

\subsection{A Hierarchy of Reinforcement Learners?}

In many reinforcement learning problems, the state space is prohibitively large or continuous, so storing all the state
values is infeasible. In most cases this problem is resolved by applying some kind of function approximation. It is
well known that neural networks have excellent function approximation capabilities, so it is no wonder that they have
been widely applied in various RL problems. These neural networks are all feed-forward ones, which is completely
satisfactory for estimating the value function of Markovian problems, as there is no need to keep past memories.

However, the application of recurrent neural networks can also be justified, if instead of value function approximation
they are used for identifying abstract states of the problem: they are able to identify \emph{spatio-temporal} regions
of the state space, not only spatial ones. This means that they give a tool for handling non-Markovian, partially
observable problems. Furthermore, their prediction capability provides a model of the environment as well.

As a consequence, RNNs with PIRANHA fit very naturally into a hierarchical RL scheme: The states of the lower level are
the raw input (which is continuous and non-Markovian). The lower-level agent uses the reconstruction/prediction error
as reward signal and is able to learn identifying some region of the input, providing high-level state description.
This state description can then be used by a traditional RL agent on the upper level. Working out the details of this
hierarchical structure is the topic of ongoing research, see, e.g.,
(\cite{szita02reinforcement,lorincz03event,szita03epsilon,bakker04hierarchical,bakker04hierarchical_b}) and references
therein.

\subsection{Summary}

In this article, a solution to the sequence replay problem was proposed. The question was to represent time series data
so that it can be reproduced sequentially from seeing only a small portion of it. We formalized the task as the
minimization of the cost function of a recurrent neural network. An analogy to reinforcement learning enabled us to
adapt policy iteration method to our problem, yielding a novel algorithm that we called PIRANHA. We showed that PIRANHA
is convergent, and fits naturally into the RL framework, yielding a hierarchical architecture.

\end{document}